\def\year{2021}\relax
\documentclass[letterpaper]{article} 
\usepackage{aaai21}  
\usepackage{times}  
\usepackage{helvet} 
\usepackage{courier}  
\usepackage[hyphens]{url}  
\usepackage{graphicx} 
\usepackage{natbib}  
\usepackage{caption} 
\usepackage{multirow}
\usepackage{amsmath}
\usepackage{algorithm}
\usepackage{algorithmic}
\usepackage{caption}
\usepackage{subfigure}

\title{LightXML: Transformer with Dynamic Negative Sampling for High-Performance Extreme Multi-label Text Classification}

\author{
Ting Jiang\textsuperscript{\rm 1}, Deqing Wang\textsuperscript{\rm 1}, Leilei Sun\textsuperscript{\rm 1,}\thanks{Corresponding Author}, Huayi Yang\textsuperscript{\rm 1}, Zhengyang Zhao\textsuperscript{\rm 1}, Fuzhen Zhuang\textsuperscript{\rm 2,3}
    
}
\affiliations{

    \textsuperscript{\rm 1}SKLSDE and \textit{BDBC} Lab, Beihang University, Beijing, China \\
    \textsuperscript{\rm 2}Key Lab of Intelligent Information Processing of CAS, Institute of Computing Technology, CAS, Beijing, China\\
    \textsuperscript{\rm 3}Beijing Advanced Innovation Center for Imaging Theory and Technology, Academy for Multidisciplinary Studies, Capital Normal University, Beijing, China \\

    \{royokong, dqwang, leileisun, yanghy, zzy979\}@buaa.edu.cn, zhuangfuzhen@ict.ac.cn\\
}

\begin{document}

\maketitle

\begin{abstract}
 Extreme Multi-label text Classification (XMC) is a task of finding the most relevant labels from a large label set. Nowadays deep learning-based methods have shown significant success in XMC. However, the existing methods (e.g., AttentionXML and X-Transformer etc) still suffer from 1) combining several models to train and predict for one dataset, and 2) sampling negative labels statically during the process of training label ranking model, which reduces both the efficiency and accuracy of the model. 
 To address the above problems, we proposed LightXML, which adopts end-to-end training and dynamic negative labels sampling. 
 In LightXML, we use generative cooperative networks to recall and rank labels, in which label recalling part generates negative and positive labels, and label ranking part distinguishes positive labels from these labels. Through these networks, negative labels are sampled dynamically during label ranking part training by feeding  with the same text representation. 
Extensive experiments show that LightXML outperforms state-of-the-art methods in five extreme multi-label datasets with much smaller model size and lower computational complexity. In particular, on the Amazon dataset with  670K labels, LightXML can reduce the model size up to 72\% compared to AttentionXML. 
Our code is available at \url{http://github.com/kongds/LightXML}.
  
\end{abstract}

\section{Introduction}
Extreme Multi-label text Classification (XMC) is a task of finding the most relevant labels for each text from an extremely large label set. It is a very practical problem that has been widely applied in many real-world scenarios, such as tagging a Wikipedia article with most relevant labels \cite{dekel2010multiclass}, dynamic search advertising in E-commerce \cite{prabhu2018parabel}, and suggesting keywords to advertisers on Amazon \cite{chang2020taming}.

Different from the classical multi-label classification problem, the candidate label set could be very large, which results in a huge computational complexity. To overcome this difficulty, many methods (e.g., Parabel \cite{prabhu2018parabel}, DiSMEC \cite{babbar2017dismec} and AttentionXML \cite{you2019attentionxml} etc) have been proposed in recent years. From the perspective of text representation learning, these methods could be divided into two categories: 1) Raw feature methods, where texts are represented by sparse vectors and feed to classifiers directly; 2) Semantic feature methods, where deep neural networks are usually employed to transfer the texts into semantic representations before the classification procedure.

The superiority of XMC methods with semantic features have been frequently reported recently. For example, AttentionXML \cite{you2019attentionxml} and X-Transformer \cite{chang2020taming} have achieved significant improvements in accuracy comparing to the state-of-the-art methods. 
However, the challenge of how to solve the XMC problem with the constraint computational resources still remains, both AttentionXML and X-Transformer need large computational resources to train or to implement. 
AttentionXML needs to train four seperate models for big XMC datasets like Amazon-670K. 
In X-Transformer, it uses a large transformer model for recalling labels, and simple linear classifications are used to rank labels.
Both methods split the training process into multiple stages, and each stage needs a separate model to train, 
which takes a lot of computational resources. Another disadvantage of these methods is the static negative label sampling. Both methods train label ranking model with negative labels sampled by fine-tuned label recalling models. This negative sampling strategy makes label ranking model only focus on a small number of negative labels, and hard to converge due to these negative labels are very similar to positive labels. 

To address the above problems, we propose a light deep learning model, LightXML, which fine-tunes single transformer model with dynamic negative label sampling. LightXML consists of three parts: text representing, label recalling, and label ranking. For text representing, we use multi-layer features of the transformer model as text representation, which can prove rich text information for the other two parts. With the advantage of dynamic negative sampling, we propose generative cooperative networks to recall and rank labels. For the label recalling part, we use the generator network based on label clusters, which is used to recall labels. For the label ranking part, we use the discriminator network to distinguish positive labels from recalled labels. 
In summary, the contributions of this paper are as follows:
\begin{itemize}
\item A novel deep learning method is proposed, which combines the powerful transformer model with generative cooperative networks. This method can fully exploit the advantage of transformer model by end-to-end training.
\item We propose dynamic negative sampling by using generative cooperative networks to recall and rank labels. The dynamic negative sampling allows label ranking part to learn from easy to hard and avoid overfitting, which can boost overall model performance.
\item Our extensive experiments show that our model achieves the best results among all methods on five benchmark datasets. In contrast, our model has much smaller model size and lower computational complexity than current state-of-the-art methods.
\end{itemize}

\section{Related work}
Many novel methods have been proposed to improve accuracy while controlling computational complexity and model sizes in XMC. These methods can be broadly categorized into two directions according to the input: One is traditional machine learning methods that use the sparse features of text like BOW features as input, and the other is deep learning methods that use raw text.
For traditional machine learning methods, we can continue to divide these methods into three directions: one-vs-all methods, tree-based methods and embedding-based methods.


\subsubsection{One-vs-all methods} One-vs-all methods such as DiSMEC \cite{babbar2017dismec}, ProXML \cite{babbar2019data}, PDSparse \cite{yen2016pd}, and PPDSparse \cite{yen2017ppdsparse} which treat each label as binary classification problem and classification tasks are independent of each other. Although many one-vs-all methods like DiSMEC and PPDSparse focus on improving model efficiency, one-vs-all methods still suffer from expensive computational complexity and large model size. With the cost of efficiency, these methods can achieve acceptable accuracy.

\subsubsection{Tree-based methods} Tree-based methods aim to overcome high computational complexity in one-vs-all methods. These methods will construct a hierarchical tree structure by partitioning labels like Parabel \cite{prabhu2018parabel} or sparse features like FastXML \cite{prabhu2014fastxml}. For Parabel, the label tree is built by label features using balance $k$-mean clustering, each node in the tree contains several classifications to classify whether the text belongs to the children node in inner nodes or the label in leaf nodes. For FastXML, FastXML directly optimizes the normalized Discounted Cumulative Gain (nDCG), and each node of FastXML contains several binary classifications to decide which children nodes to traverse like Parabel.

\subsubsection{Embedding-based methods}Embedding-based methods project high dimensional label space into a low dimensional space to simplify the XMC problem. 
And the design of label compression part and label decompression part is significant for the performance of these methods. However, no matter how the label compression part is designed, label compression will always lose a part of information. It makes these methods achieve worse accuray compared with one-vs-all methods and methods.
And some improved embedding-based methods like SLEEC \cite{bhatia2015sparse} and AnnexML \cite{tagami2017annexml} be proposed to solve this problem by improving the label compression and decompression parts, but the problem still remains.

\subsubsection{Deep learning methods}
With the development of NLP, deep learning methods have shown great improvement in XMC which can learn better text representation from raw text. But the main challenge to these methods is how to couple with millions of labels with limited GPU resources. we review three most representative methods, e.g., XML-CNN \cite{liu2017deep}, AttentionXML \cite{you2019attentionxml} and X-Transformers \cite{chang2020taming}..

XML-CNN is the first successful method that showed the power of deep learning in XMC. It learns text representation by feeding word embedding to CNN networks with end-to-end training. For scaling to the datasets with hundreds of thousands of labels, XML-CNN proposes a hidden bottleneck layer to project text feature into low dimensional space, which can reduce the overall model size. But XML-CNN only uses a simple fully connected layer to score all labels with binary entropy loss like simple multi-label classification, which makes it hard to deal with large label sets.

After XML-CNN, AttentionXML shows great success in XMC, which overpassed all traditional machine learning methods and proved the superiority of the raw text compare to sparse features. Unlike using a simple fully connected layer for label scoring in XML-CNN, AttentionXML adopts a probabilistic label tree (PLT) that can handle millions of labels. In AttentionXML, it uses RNN networks and attention mechanisms to handle raw text with diffferent models to each layer of PLT. It needs to train several models for one dataset. For solving this problem, AttentionXML initializes the weight of the current layer model by its upper layer model, which can help the model converge quickly. But It still makes AttentionXML much slow in predicting and surffers from big overall model size. In conclusion, AttentionXML is an enlightened method that elegantly combines PLT with deep learning methods.

For X-Transformer, 
it only uses deep learning models to match the label clusters for given raw text and ranks these labels by high dimension linear classifications with the sparse feature and text representation of deep learning models. 
X-Transformer is the first method of using deep transformer models in XMC. Due to the high computational complexity of transformer models, it only fine-tunes transformer models as the label clusters matcher, which can not fully exploit the power of transformer models. 
Although X-Transformer can reach higher accuracy than AttentionXML with the cost of high computational complexity and model size, which makes X-Transformer infeasible in XMC applications. Because AttentionXML can reach better accuracy with the same computational complexity of X-Transformer by using more models for ensemble.

\begin{figure}[h]
\centering
\includegraphics[width=0.95\columnwidth]{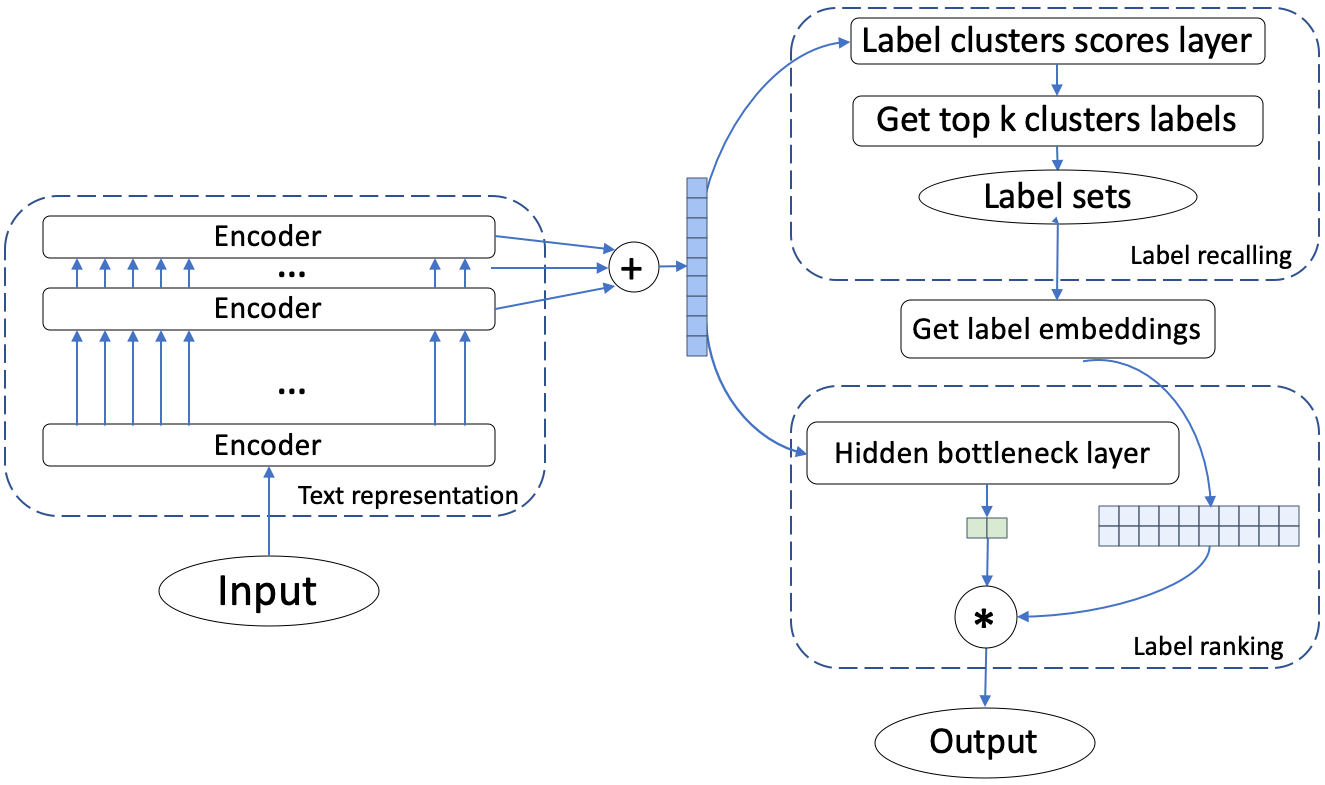} 
\caption{An overview of the proposed framework}.
\label{pic:framework}
\end{figure}

\section{Methodology}
\subsection{Problem Formulation}
Given a training set \(\left\{(\mathbf{x}_i, \mathbf{y}_i)\right\}^N_{i=1}\) where \(\mathbf{x}_i\) is raw text, and \(\mathbf{y}_i \in \left\{0, 1\right\}^L\) is the label of \(\mathbf{x}_i\) represented by $L$ dimensional multi-hot vectors. Our goal is to learn a function $f(\mathbf{x}_i) \in R^L $ which gives scores to all labels, and $f$ needs to give high score to $l$ with \(\mathbf{y}_{il}=1\) for $\mathbf{x}_i$. We can obtain top-K predicted labels by $f(\mathbf{x}_i)$. But in many XMC methods, it only scores the recalled labels to reduce the overall computational complexity, and the number of these labels in the subset will be much smaller than all labels, which can save much computation time.
\subsection{Framework}

The proposed framework shows in Figure \ref{pic:framework}. We first cluster labels by sparse features of labels. After label clustering, we have a certain number of label clusters, and each label belongs to one label cluster. Then we  employ the transformer model to embed raw text information into a high dimension representation, which is the input of label recalling part and label ranking part.

For label recalling part and label ranking part, with the advantage of dynamic negative sampling, we proposed generative cooperative networks for XMC. The label recalling part is the generator of these networks, which can dynamically sample negative labels. The label ranking part is the discriminator of these networks, which distinguishes between the negative labels and positive labels. 
For the cooperation of these networks, the generator assists the discriminator learning better label representation, and the discriminator helps the generator to consider fine-grained text information. Because the discriminator is trained according to specific label rather than label clustering.

Specifically, for the generator, we score every label cluster by text representation to obtain top-K label clusters which are the labels that we sampled. All positive labels are added to this subset in the training stage to make the discriminator can be fine-tuned with generator. For the discriminator, each label in this label subset will be scored to distinguish the positive labels and the negative labels.
\subsubsection{Label clustering}
Label clustering is equivalent to two layers of Probabilistic Label Tree (PLT). Since the deep PLT will harm performance and label clustering is enough to cope with extreme numbers of labels, we just use this two-layer PLT with the same constructing methods in AttentionXML \cite{you2019attentionxml}.

More specifically, given a maximum number of labels that each cluster $s$ has, our goal is to partition labels into $K$ label clusters, and the number of labels contained in each label cluster is less than $s$ and greater than $s/2$. To solve this problem, We first get each label representation by normalizing the sum of sparse text features with corresponding labels contain this label. Then, we use balanced k-means (k=2) clustering to recursively partition label sets until all label sets satisfy the above requirement.
\subsubsection{Text representation}
Transformer models show outstanding performance on a wide array of NLP tasks. In our framework, We adopt three pre-trained transformer base models: BERT \cite{devlin2018bert}, XLNet \cite{yang2019xlnet}  and RoBERTa \cite{liu2019roberta}, which is the same as X-Transformer \cite{chang2020taming}. But compared to large transformer models(24 layers and 1024 hidden dimension) that X-Transformer uses, we only use base transformer models (12 layers and 768 hidden dimension) to reduce computational complexity.

For input sequence length, the time and space complexity of transformer models will grow exponentially with the growth of text length under self-attention mechanisms \cite{vaswani2017attention}, which makes transformer models hard to deal with long text. In X-Transformer, maximum sequence length is set to 128 for all datasets. However, we set maximum sequence length to 512 for small XMC datasets and 128 for large XMC datasets with the advantage of using base models instead of large models.

For text embedding of transformer models, different from fine-tuning transformer models for typical text classification tasks. In order to make full use of the transformer models in XMC, we concatenate the "[CLS]" token in the hidden state of last five layers as text representation. Let \(e_i \in R^l\) be the $l$ dimensional hidden state of the "[CLS]" token in the last $i$-th layer of the transformer model. The text is represented by the concatenation of last five layers \( e=\left[e_{1}, . ., e_{5}\right] \in R^{5 l} \). High dimensional text representation can enrich text information and improve the generalization ability of overall model in XMC. And our experiments show it can speed up convergence and improve the model performance. To avoid overfitting, We also use a high rate of dropout to this high dimensional text representation

\subsubsection{Label recalling}
In this part, our goal is not only sampling positive labels, but also negative labels to help the label ranking part learn. To reduce computational complexity and speed up convergence, we directly sample the label clusters instead of the labels, and use all labels in these clusters as the labels we sample.

The generator is a fully connected layer with sigmoid \( G(e)=\sigma(W_g e + b_g) \) and $G$ returns $K$ dimensional vector representation, which is the scores of all $K$ label clusters. We choose top $b$ label clusters to generate a subset of labels. 
In training, all positive labels are added to this subset to force teaching the label ranking to distinguish positive and negative labels. In predicting, we don't modify this subset, and this subset may not contain all positive labels.

For the generator loss, we don't calculate loss according to the feedback of the discriminator, due to the generator is designed to cooperate with the discriminator and make model easy to convergence. And we can directly calculate the loss by the ground truth. The loss function is described as follows:
\begin{equation}\label{eq:G_loss}
\begin{split}
    \mathcal{L}_g(G(e), y_g) = &\sum_{i=0}^{K}(1-y_g^i)(-log(1-G(e)^i)) +\\
    &y_g^i(-log(G(e)^i)),
\end{split}
\end{equation}
where \(y_g \in \{0, 1\}^{K}\) is the multi-hot representation of label cluster for the given text.

 Negative label sampling is a decisive factor for overall model performance. In AttentionXML \cite{you2019attentionxml}, negative samples are static, and the model only overfits to distinguish specific negative label samples, which will constrain the performance. The static negative sampling also makes the model hard to converge, because negative labels are very similar to positive labels. We solve this problem by employing the generator with dynamically negative labels sampling. For the same training instance, the negative labels of this instance is resampled every time by the current generator, and negative labels are sampled from easy to difficult distinguishing during the generator fiting, which makes the discriminator converge easily and avoid overfitting.
 The label candidates the generator sampled are as follows:
\begin{equation}
    S_g = \left\{l_i : i \in \left\{i : g_c(l_i) \in G(e) \right\} \right\},
\end{equation}
where \(g_c\) is a function to map labels to its clusters and \(l_i\) is the  $i$-th label. In training stage, all positive labels are added to $S_g$.

\subsubsection{Label ranking}
Given text representation $e$ and label candidates $S_g$, we first need to get embeddings of all labels in $S_g$ which can represent as follows:
\begin{equation}
     M = \left[E_i : i \in \left\{i : S_g \right\} \right],
\end{equation}
where \(E_i \in R^b\) is the learn-able $b$ dimension embedding of $i$-th label and \(E \in R^{L \times b}\) is the overall label embedding matrix which is initialized randomly.

We use the same hidden bottleneck layer as XML-CNN \cite{liu2017deep} to project text embedding to low dimension. There are two advantages of it: 
\begin{itemize}
\item The hidden bottleneck layer makes the overall model size smaller and let the model fit into limited GPU memory. The overall label size $L$ is usually more than hundreds of thousands in XMC. If we remove this layer, this part is size will be \(O(L \times 5k)\), which can take huge GPU memory. After adding a hidden bottleneck layer, the size will be \(O((L + 5k) \times b)\), and the hyper-parameter $b$ is the dimension of the label embedding, which is much smaller than $5k$. According to the size of different datasets, we can set different $b$ to make full use of GPU memory. 
\item The hidden bottleneck layer also makes the generator and the discriminator focus on different information of text representation. The generator focuses on fine-grained text information, while the discriminator focuses on coarse-grained text information. And the final results will combine these two types of information.
\end{itemize}

The discriminator can be described as follows:
\begin{equation}
D(e, M) = \sigma(M \sigma(W_h e + b_h)),
\end{equation}
where \( W_h \in R^{b \times 5k} \) and \( b_h \in R^b \) is the weight of hidden bottleneck layer.

The object of $D(e, M)$ is to distinguish positive labels and negative labels that are sampled by the generator. The training target of $D(e, M)$ is $y_d$ where $y_d^i = 0$ if $S_g^i$ is positive labels and $y_d^i = 1$ if $S_g^i$ is negative label. Thus the loss of this part is as follows:
\begin{equation}\label{eq:D_loss}
\begin{split}
    \mathcal{L}_d(D(e, M), y_d) = &\sum_{i=0}^{K}(1-y_d^i)(-log(1-D(e, M)^i)) +\\
    &y_d^i(-log(D(e, M)^i))
\end{split}
\end{equation}

The whole framework of LightXML is shown in Algorithm 1.
\renewcommand{\algorithmicrequire}{\textbf{Input:}}
\renewcommand{\algorithmicensure}{\textbf{Output:}}

\begin{algorithm}[htb]
  \caption{ The proposed framework. }  \label{alg:Framework}
  \begin{algorithmic}[1]
    \REQUIRE Training set \(\left\{X, Y \right\} = \left\{(\mathbf{x}_i, \mathbf{y}_i)\right\}^N_{i=1}\), sparse feature of training text $\hat{X}$;
    \ENSURE Model;
      \STATE Construct the label clusters $C$ by $\hat{X}$ and $Y$;
      \STATE Initialize transformer model $T$ with pre-trained transformer model;
      \STATE Initialize discriminator $D$ base on $C$;
      \STATE Initialize label embedding $E$, generator $G$;
      \WHILE {model not converge}
        \STATE Draw $m$ samples $X_{batch}$ and $Y_{batch}$ from the training set \(\left\{X, Y \right\}\);
        \STATE Get text embedding $Z$ by $T(X_{batch})$;
        \FOR{i=1..$m$}
            \STATE Generate label clusters $S_{generated}$ by \(G(Z_i)\);
            \STATE Get negative labels $S_{neg}$ by $S_{generated}$ and $C$;
            \STATE Remove positive labels in $S_{neg}$;
        \ENDFOR
        \STATE Get positive labels $S_{pos}$ according to $Y_{batch}$;
        \FOR{i=1..$m$}
            \STATE Get label embedding $M$ according to $S_{pos}$ and $S_{neg}$;
            \STATE Generate each label scores by \(D(Z_i, M)\);
        \ENDFOR
        \STATE Update parameters of $T$, $G$ and $D$ according to Eq.\ref{eq:G_loss},  and Eq.\ref{eq:D_loss};
      \ENDWHILE
    \RETURN model;
  \end{algorithmic}
\end{algorithm}

\subsection{Training}
Unlike AttentionXML \cite{you2019attentionxml} and X-Transformer \cite{chang2020taming}, our model can perform end-to-end training by using generative cooperative networks to handle both label recalling and label ranking parts, which can reduce training time and model size. The overall loss function is:
\begin{equation}
    \mathcal{L} = \mathcal{L}_g + \mathcal{L}_d
\end{equation}
We directly add the loss of generator part $\mathcal{L}_g$ and discriminator part $\mathcal{L}_d$ as overall loss, and the transformer model that used to represent text will update its gradient according to both $\mathcal{L}_g$ and $\mathcal{L}_d$, which makes the transformer model learn from both label recalling and label ranking.

\subsection{Prediction}
The efficiency of prediction is essential for XMC applications. But deep learning methods reach high accuracy with the huge cost of computational complexity, which makes them infeasible compared to traditional machine learning methods. Although LightXML is based on deep transformer models, LightXML can make fast prediction in several milliseconds on the large scale XMC dataset. 

LightXML can also end-to-end predict with raw text. The label recalling part of LightXML scores all label clusters and return a subset of all labels containing both positive labels and negative labels. The label ranking part scores every label in this subset. The final scores of labels are the multiplying of recalling scores and ranking scores, and we can get top-K labels by these scores.

\section{Experiments}
\subsection{Dataset}
Five widely-used XMC benchmark datasets are used in our experiments, they are: Eurlex-4K \cite{mencia2008efficient}, Wiki10-31K \cite{zubiaga2012enhancing}, AmazonCat-13K \cite{mcauley2013hidden}, Wiki-500K and Amazon-670K \cite{mcauley2013hidden}. The detailed information of each dataset is shown in Table \ref{datasets detail information}. 
The sparse feature of text we use for clustering is also contained in datasets.
\subsection{Evalutaion Measures}
We choose $\mathrm{P} @ k$ as evaluation metrics, which is widely used in XMC and represents percentage of accuracy labels in top $k$ score labels.$\mathrm{P} @ k$ can be defined as follows:
\begin{equation}
    P @ k=\frac{1}{k} \sum_{i \in \operatorname{rank}_{k}(\hat{y})} y_{i},
\end{equation}
where \( \hat{y} \) is the prediction vector, $i$ denotes the index of the i-th highest element in \( \hat{y} \) and $y \in\{0,1\}^{L}$.

\subsection{Baseline}
We compared the state-of-the-art and most enlightening methods including one-vs-all DiSMEC \cite{babbar2017dismec}; label tree based Parabel \cite{prabhu2018parabel}, ExtremeText (XT) \cite{wydmuch2018no}, Bonsai \cite{khandagale2019bonsai}; and deep learning based XML-CNN \cite{liu2017deep}, AttentionXML \cite{you2019attentionxml}, and X-Transformers \cite{chang2020taming}. The results of all baseline are from \cite{you2019attentionxml} and \cite{chang2020taming}.

\subsection{Experiment Settings}
For all datasets, we directly use raw texts without preprocessing, and texts are truncated to the maximum input tokens. We use 0.5 rate of dropout for text representation and SWA (stochastic weight averaging) \cite{izmailov2018averaging} which is also used in AttentionXML to avoid overfitting. LightXML is trained by AdamW \cite{kingma2014adam} with constant learning rate of 1e-4 and 0.01 weight decay for bias and layer norm weight in model.  Automatic Mixed Precision (AMP) is also used to reduce GPU memory usage and increase training. Our training stage is end-to-end, and the loss of model is the sum of label recalling loss and label ranking loss. For datasets with small labels like Eurlex-4k, Amazoncat-13k and Wiki10-31k, each label clusters contain only one label and we can get each label scores in label recalling part. 
For ensemble, we use three different transformer models for Eurlex-4K, Amazoncat-13K and Wiki10-31K, and use three different label clusters with BERT \cite{devlin2018bert} for Wiki-500K and Amazon-670K.
Compared to state-of-the-art deep learning methods, all of our models is trained on single Tesla v100 GPU, and our model only uses less than 16GB of GPU memory for training, which is much smaller than other methods use. Other hyperparameters is given in Table \ref{tab:hyperparameter}.
\begin{table}[h]
\centering
\begin{tabular}{c|ccccc}
\hline
Datasets & $E$ & $B$ & $b$ & $C$ & $L_t$ \\
\hline
Eurlex-4K & 20 & 16 & - & - & 512 \\
AmazonCat-13K & 5 & 16 & - & - & 512 \\
Wiki-31K & 30 & 16 & - & - & 512 \\
Wiki-500K & 10 & 32 & 500 & 60 & 128 \\
Amazon-670K & 15 & 16 & 400 & 80 & 128 \\
\hline
\end{tabular}
\caption{Hyperparameters of all datasets.
        \(E\) is the number of epochs, \(B\) is the batch size, \( b \) is the dimension of label embedding, \( C \) is the number of labels in one label cluster, \( L_t \) is the maximum length of transformer model is input tokens and \(M\) is the model size.
} \label{tab:hyperparameter}
\end{table}

\begin{table*}[h]
\centering
\begin{tabular}{crrrrrrrr}
\hline
Datasets & \(N_{train}\) & \(N_{test}\) & $D$ & $L$ & \(\bar{L}\) & \(\hat{L}\) & \(\bar{W}_{train}\) & \(\bar{W}_{test}\) \\
\hline
Eurlex-4K & 15,449 & 3,865 & 186,104 & 3,956 & 5.30 & 20.79 & 1248.58 & 1230.40 \\
Wikil0-31K & 14,146 & 6,616 & 101,938 & 30,938 & 18.64 & 8.52 & 2484.30 & 2425.45 \\
AmazonCat-13K  & 1,186,239 & 306,782 & 203,882 & 13,330 & 5.04 & 448.57 & 246.61 & 245.98\\
Amazon-670K & 490,449 & 153,025 & 135,909 & 670,091 & 5.45 & 3.99 & 247.33 & 241.22\\
Wiki-500K & 1,779,881 & 769,421 & 2,381,304 & 501,008 & 4.75 & 16.86 & 808.66 & 808.56\\
\hline
\end{tabular}
  \caption{Detailed datasets statistics.
        \(N_{train}\) is the number of training samples, \(N_{test}\) is the number of test samples, $D$ is the dimension of BOW feature vector, $L$ is the number of labels, $\bar{L}$ is the average number of labels per sample, $\hat{L}$ is the average number of samples per label, $\bar{W}_{train}$ is the average number of words per training sample and $\bar{W}_{test}$ is the average number of words per testing sample.
  } \label{datasets detail information}
\end{table*}
\begin{table*}[h]
\centering
\small
\begin{tabular}{cccccccccc}
\hline
 Datasets & & DiSMEC & Parabel & Bonsai & XT & XML-CNN & AttentionXML & X-Transformer &  LightXML\\
\hline
\multirow{3}*{Eurlex-4K} & $P@1$ & 83.21 & 82.12 & 82.30 & 79.17 & 75.32 & 87.12 & 87.22 & \textbf{87.63} \\
~ & $P@3$ & 70.39 & 68.91 & 69.55 & 66.80 & 60.14 & 73.99 & 75.12 & \textbf{75.89} \\
~ & $P@5$ & 58.73 & 57.89 & 58.35 & 56.09 & 49.21 & 61.92 & 62.90 & \textbf{63.36} \\
\hline
\multirow{3}*{AmazonCat-13K} & $P@1$ & 93.81 & 93.02 & 92.98 & 92.50 & 93.26 & 95.92 & 96.70 & \textbf{96.77} \\
~ & $P@3$ & 79.08 & 79.14 & 79.13 & 78.12 & 77.06 & 82.41 & 83.85 & \textbf{84.02} \\
~ & $P@5$ & 64.06 & 64.51 & 64.46 & 63.51 & 61.40 & 67.31 & 68.58 & \textbf{68.70} \\
\hline
\multirow{3}*{Wiki10-31K} & $P@1$ & 84.13 & 84.19 & 84.52 & 83.66 & 81.41 & 87.47 & 88.51 & \textbf{89.45} \\
~ & $P@3$ & 74.72 & 72.46 & 73.76 & 73.28 & 66.23 & 78.48 & 78.71 & \textbf{78.96} \\
~ & $P@5$ & 65.94 & 63.37 & 64.69 & 64.51 & 56.11 & 69.37 & 69.62 & \textbf{69.85} \\
\hline
\multirow{3}*{Wiki-500K} & $P@1$ & 70.21 & 68.70 & 69.26 & 65.17 & - & 76.95 & 77.28 & \textbf{77.78}\\
~ & $P@3$ & 50.57 & 49.57 & 49.80 & 46.32 & - & 58.42 & 57.47 & \textbf{58.85} \\
~ & $P@5$ & 39.68 & 38.64 & 38.83 & 36.15 & - & \textbf{46.14} & 45.31 & 45.57 \\
\hline
\multirow{3}*{Amazon-670K} & $P@1$ & 44.78 & 44.91 & 45.58 & 42.54 & 33.41 & 47.58 & - & \textbf{49.10} \\
~ & $P@3$ & 39.72 & 39.77 & 40.39 & 37.93 & 30.00 & 42.61 & - & \textbf{43.83} \\
~ & $P@5$ & 36.17 & 35.98 & 36.60 & 34.63 & 27.42 & 38.92 & - & \textbf{39.85} \\
\hline
\end{tabular}
\caption{Comparisons with different methods.
       Comparing our model against state-of-the-art XMC methods on Eurlex-4K, AmazonCat-13K, Wiki10-31K, Wiki-500K and Amazon-670K. Note that XML-CNN are not scalable on Wiki-500K, and the result of X-Transformer on Amazon-670K has never been reported which is hard to reproduce it limited by our hardware conditions.
} \label{tab:performance}
\end{table*}
\subsection{Performance comparison}
Table \ref{tab:performance} shows $P@k$ on the five datasets. We focus on top prediction by varying $k$ at 1, 3 and 5 in $P@K$ which are widely used in XMC. LightXML outperforms all methods on four datasets. For traditional machine learning methods, DiSMEC has better accuracy compared to these methods (Parabel, Bonsai and XT) with the cost of high computational complexity, and LightXML has much improvement on accuracy compared to these methods. For X-Transformer, which is also transformer models based method, LightXML achieve better accuracy on all datasets with much small model size and computational complexity, which can prove the effectiveness of our method. For AttentionXML, although AttentionXML has slightly better $P@5$ 
than LightXML on Wiki-500K, but LightXML achieves more improvement in $P@1$ and $P@3$, and LightXML outperforms AttentionXML on other four datasets.

\subsection{Performance on single model}
We also examine the single model performance, called LightXML-1. Table \ref{single performance} shows the results of single models on Amazon-670K and Wiki-500K. LightXML-1 shows better accuracy compare to AttentionXML-1. 

\begin{table}[h]
\centering
\begin{tabular}{cccc}
\hline 
Datasets&  & LightXML-1 & AttentionXML-1\\
 \hline
\multirow{3}*{Wiki-500K} & $P@1$ & 76.19 & 75.07 \\
~ & $P@3$ & 57.22 & 56.49 \\
~ & $P@5$ & 44.12 & 44.41 \\
 \hline
\multirow{3}*{Amazon-670K} & $P@1$ & 47.14 & 45.66 \\
~ & $P@3$ & 42.02 & 40.67 \\
~ & $P@5$ & 38.23 & 36.94 \\
\hline
\end{tabular}
\caption{Performances on single model} \label{single performance}
\end{table}

\subsection{Effect of the dynamic negative sampling}
To examine the importance of the dynamic negative sampling in negative label sampling, we compare dynamic negative sampling with static negative sampling on Wiki-500K and Amazon-670K.
 



\begin{table}[h]
\centering
\begin{tabular}{ccccc}
\hline 
Datasets&  & $D$ & $BS$ & $S$\\
 \hline
\multirow{3}*{Wiki-500K} & $P@1$ & 76.19 & 75.30 & 73.30 \\
~ & $P@3$ & 57.22 & 56.64  & 54.60 \\
~ & $P@5$ & 44.12 & 43.03  & 42.27 \\
 \hline
\multirow{3}*{Amazon-670K} & $P@1$ & 47.14 & 46.27 & 43.36 \\
~ & $P@3$ & 42.02 & 41.40 & 38.55 \\
~ & $P@5$ & 38.23 & 37.69 & 34.96 \\
\hline
\end{tabular}
\caption{Comparisons with static negative sampling methods. $D$ is the dynamic negative sampling, $BS$ is the static negative sampling with additional text representation and $S$ is the static negative sampling with single text representation. 
} \label{tab:static_negative}

\end{table}
For static negative sampling, it's hard to have a fair comparison with dynamic negative sampling, due to the constraint of static negative sampling, which needs to train recalling and ranking in order. So we proposed two version of static negative sampling: 1) $S$ has the same model size with dynamic negative sampling, and we trained label recalling with the freeze text representation of trained label ranking. 2) $BS$ has a additional text representation, which can been fine-tuned in label ranking. And we initialize this with the text representation of trained label recalling.

The performance of different negative sampling is shown in Table \ref{tab:static_negative}. Although both two static negative sampling methods take longer than dynamic negative sampling methods in training time, dynamic negative sampling method still outperforms the other two methods. For two static negative sampling methods, $BS$ shows better results than $S$ with additional text representation.

\subsection{Effect of the multi layers text representation}
This section analyze how multi layers text representation affect the model performance, and we choose our model and our model with only single last layer of "[CLS]" token as text representation. 

\begin{figure}[h]
\centering    

\subfigure[Wiki-500K] 
{
	\includegraphics[width=0.92\columnwidth]{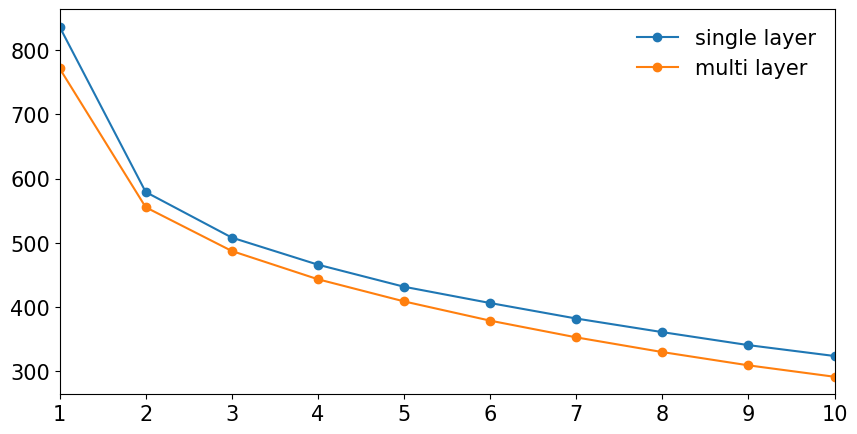}   
}
\subfigure[Amazon-670K] 
{
	\includegraphics[width=0.92\columnwidth]{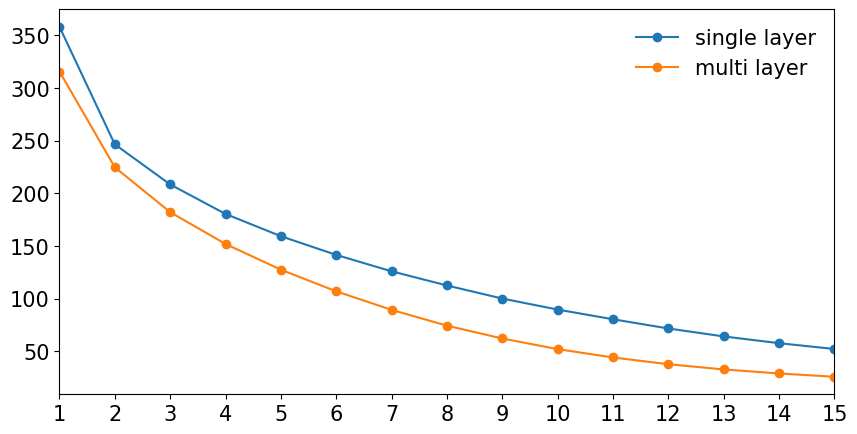}   
}
\caption{Effect of the multi layers text representation.
} 
\label{fig:ab2}
\end{figure}

As Figure \ref{fig:ab2} shows, we compare the training loss of multi layers and single layer on Wiki-500K and Amazon-670K. It can be seen that multi layers has low training loss, which means multi layers text representation can accelerate model convergence, and multi layers can reach same training loss of single layer by only using half of total epochs. For final accuracy, multi layers can improve the final accuracy of $P@5$ more than 1\%. 

\subsection{Computation Time and Model Size}
 Computation time and model size are essential for XMC, which means XMC is not only a task to pursue high accuracy, but also a task to improve efficiency.
In this section, we compare the computation time and model size of LightXML with high performance XMC method AttentionXML. 
For X-Transformer, it uses large transformer models and high dimension linear classification, which makes it has large model size and high computational complexity. X-Transformer takes more than 35 hours of training with eight Tesla V100 GPUs on Wiki-500K, and it will need more than one hundred hours for us to reproduce. Due to the hard reproducing and bad efficiency of the X-Transformer, we don't compare LightXML with X-Transformer.

\begin{table}[h]
    \centering
    \small
    \begin{tabular}{ccccc}
        \hline
         Datasets & &  AttentionXML-1 &  LightXML-1  \\ 
         \hline
         \multirow{3}*{Wiki-500K} & $T_{train}$ & 37.28 & \textbf{34.33} \\
         ~ & $S_{test}$ & 4.20 & \textbf{3.75} \\
         ~ & $M$ & 3.11 & \textbf{1.47}\\
         \hline
         \multirow{3}*{Amazon-670K} &   $T_{train}$ & \textbf{26.10} & 28.75 \\
         ~ & $S_{test}$ & 8.59 & \textbf{4.09} \\
         ~ & $M$ & 5.52 & \textbf{1.53}\\
         \hline
    \end{tabular}
    \caption{Computation Time and Model Size.
    $T_{train}$ is the overall training hours.
    $S_{test}$ is the average time required to predict each sample. The unit of $S_{test}$ is milliseconds per sample (ms/sample). $M$ is the model size in GB.
    }\label{tab:computation_time}
\end{table}
Tabel \ref{tab:computation_time} shows the training time, predicting speed and model size of AttentionXML-1 and LightXML-1 on Wiki-500K and Amazon-670K, and both AttentionXML and LightXML uses same hardware with one Tesla V100 GPU. LightXML shows significant improvements on both predicting speed and model size compare to AttentionXML. For predicting speed, LightXML can find relevant labels from more than 0.5 million labels in 5 milliseconds with raw text as input. For model size, LightXML can reduce 72\% model size in Amazon-670K, and 52\% on Wiki-500K. For training time, both LightXML and AttentionXML are fast in training, which can save more than three times training time compare to X-Transformer.
\section{Conclusion}
In this paper, we proposed a light deep learning model for XMC, LightXML, which combines the transformer model with generative cooperative networks. With generative cooperative networks, the transformer model can be end-to-end fine-tuned in XMC, which makes the transformer model learn powerful text representation. To make LightXMl robust in predicting, we also proposed dynamic negative sampling based on these generative cooperative networks. With extensive experiments, LightXML shows high efficiency on large scale datasets with the best accuracy comparing to the current state-of-the-art methods, which can allow all of our experiments to be performed on the single GPU card with a reasonable time.  
Furthermore, current state-of-the-art deep learning methods have many redundant parameters, which will harm the performance, and LightXML can remain accuracy  while reducing more than 50\% model size than these methods.

\section{Acknowledgment}
This work was supported by the National Natural Science Foundation of China under Grant Nos. 71901011, U1836206, and National Key R\&D Program of China under Grant No. 2019YFA0707204
\bibliography{ref} 

\begin{thebibliography}{23}
\providecommand{\natexlab}[1]{#1}
\providecommand{\url}[1]{\texttt{#1}}
\providecommand{\urlprefix}{URL }
\expandafter\ifx\csname urlstyle\endcsname\relax
  \providecommand{\doi}[1]{doi:\discretionary{}{}{}#1}\else
  \providecommand{\doi}{doi:\discretionary{}{}{}\begingroup
  \urlstyle{rm}\Url}\fi

\bibitem[{Babbar and Sch{\"o}lkopf(2017)}]{babbar2017dismec}
Babbar, R.; and Sch{\"o}lkopf, B. 2017.
\newblock Dismec: Distributed sparse machines for extreme multi-label
  classification.
\newblock In \emph{Proceedings of the Tenth ACM International Conference on Web
  Search and Data Mining}, 721--729.

\bibitem[{Babbar and Sch{\"o}lkopf(2019)}]{babbar2019data}
Babbar, R.; and Sch{\"o}lkopf, B. 2019.
\newblock Data scarcity, robustness and extreme multi-label classification.
\newblock \emph{Machine Learning} 108(8-9): 1329--1351.

\bibitem[{Bhatia et~al.(2015)Bhatia, Jain, Kar, Varma, and
  Jain}]{bhatia2015sparse}
Bhatia, K.; Jain, H.; Kar, P.; Varma, M.; and Jain, P. 2015.
\newblock Sparse local embeddings for extreme multi-label classification.
\newblock In \emph{Advances in neural information processing systems},
  730--738.

\bibitem[{Chang et~al.(2020)Chang, Yu, Zhong, Yang, and
  Dhillon}]{chang2020taming}
Chang, W.-C.; Yu, H.-F.; Zhong, K.; Yang, Y.; and Dhillon, I.~S. 2020.
\newblock Taming Pretrained Transformers for Extreme Multi-label Text
  Classification.
\newblock In \emph{Proceedings of the 26th ACM SIGKDD International Conference
  on Knowledge Discovery \& Data Mining}, 3163--3171.

\bibitem[{Dekel and Shamir(2010)}]{dekel2010multiclass}
Dekel, O.; and Shamir, O. 2010.
\newblock Multiclass-multilabel classification with more classes than examples.
\newblock In \emph{Proceedings of the Thirteenth International Conference on
  Artificial Intelligence and Statistics}, 137--144.

\bibitem[{Devlin et~al.(2018)Devlin, Chang, Lee, and
  Toutanova}]{devlin2018bert}
Devlin, J.; Chang, M.-W.; Lee, K.; and Toutanova, K. 2018.
\newblock Bert: Pre-training of deep bidirectional transformers for language
  understanding.
\newblock \emph{arXiv preprint arXiv:1810.04805} .

\bibitem[{Izmailov et~al.(2018)Izmailov, Podoprikhin, Garipov, Vetrov, and
  Wilson}]{izmailov2018averaging}
Izmailov, P.; Podoprikhin, D.; Garipov, T.; Vetrov, D.; and Wilson, A.~G. 2018.
\newblock Averaging weights leads to wider optima and better generalization.
\newblock \emph{arXiv preprint arXiv:1803.05407} .

\bibitem[{Khandagale, Xiao, and Babbar(2019)}]{khandagale2019bonsai}
Khandagale, S.; Xiao, H.; and Babbar, R. 2019.
\newblock Bonsai--Diverse and Shallow Trees for Extreme Multi-label
  Classification.
\newblock \emph{arXiv preprint arXiv:1904.08249} .

\bibitem[{Kingma and Ba(2014)}]{kingma2014adam}
Kingma, D.~P.; and Ba, J. 2014.
\newblock Adam: A method for stochastic optimization.
\newblock \emph{arXiv preprint arXiv:1412.6980} .

\bibitem[{Liu et~al.(2017)Liu, Chang, Wu, and Yang}]{liu2017deep}
Liu, J.; Chang, W.-C.; Wu, Y.; and Yang, Y. 2017.
\newblock Deep learning for extreme multi-label text classification.
\newblock In \emph{Proceedings of the 40th International ACM SIGIR Conference
  on Research and Development in Information Retrieval}, 115--124.

\bibitem[{Liu et~al.(2019)Liu, Ott, Goyal, Du, Joshi, Chen, Levy, Lewis,
  Zettlemoyer, and Stoyanov}]{liu2019roberta}
Liu, Y.; Ott, M.; Goyal, N.; Du, J.; Joshi, M.; Chen, D.; Levy, O.; Lewis, M.;
  Zettlemoyer, L.; and Stoyanov, V. 2019.
\newblock Roberta: A robustly optimized bert pretraining approach.
\newblock \emph{arXiv preprint arXiv:1907.11692} .

\bibitem[{McAuley and Leskovec(2013)}]{mcauley2013hidden}
McAuley, J.; and Leskovec, J. 2013.
\newblock Hidden factors and hidden topics: understanding rating dimensions
  with review text.
\newblock In \emph{Proceedings of the 7th ACM conference on Recommender
  systems}, 165--172.

\bibitem[{Mencia and F{\"u}rnkranz(2008)}]{mencia2008efficient}
Mencia, E.~L.; and F{\"u}rnkranz, J. 2008.
\newblock Efficient pairwise multilabel classification for large-scale problems
  in the legal domain.
\newblock In \emph{Joint European Conference on Machine Learning and Knowledge
  Discovery in Databases}, 50--65. Springer.

\bibitem[{Prabhu et~al.(2018)Prabhu, Kag, Harsola, Agrawal, and
  Varma}]{prabhu2018parabel}
Prabhu, Y.; Kag, A.; Harsola, S.; Agrawal, R.; and Varma, M. 2018.
\newblock Parabel: Partitioned label trees for extreme classification with
  application to dynamic search advertising.
\newblock In \emph{Proceedings of the 2018 World Wide Web Conference},
  993--1002.

\bibitem[{Prabhu and Varma(2014)}]{prabhu2014fastxml}
Prabhu, Y.; and Varma, M. 2014.
\newblock Fastxml: A fast, accurate and stable tree-classifier for extreme
  multi-label learning.
\newblock In \emph{Proceedings of the 20th ACM SIGKDD international conference
  on Knowledge discovery and data mining}, 263--272.

\bibitem[{Tagami(2017)}]{tagami2017annexml}
Tagami, Y. 2017.
\newblock Annexml: Approximate nearest neighbor search for extreme multi-label
  classification.
\newblock In \emph{Proceedings of the 23rd ACM SIGKDD international conference
  on knowledge discovery and data mining}, 455--464.

\bibitem[{Vaswani et~al.(2017)Vaswani, Shazeer, Parmar, Uszkoreit, Jones,
  Gomez, Kaiser, and Polosukhin}]{vaswani2017attention}
Vaswani, A.; Shazeer, N.; Parmar, N.; Uszkoreit, J.; Jones, L.; Gomez, A.~N.;
  Kaiser, {\L}.; and Polosukhin, I. 2017.
\newblock Attention is all you need.
\newblock In \emph{Advances in neural information processing systems},
  5998--6008.

\bibitem[{Wydmuch et~al.(2018)Wydmuch, Jasinska, Kuznetsov, Busa-Fekete, and
  Dembczynski}]{wydmuch2018no}
Wydmuch, M.; Jasinska, K.; Kuznetsov, M.; Busa-Fekete, R.; and Dembczynski, K.
  2018.
\newblock A no-regret generalization of hierarchical softmax to extreme
  multi-label classification.
\newblock In \emph{Advances in Neural Information Processing Systems},
  6355--6366.

\bibitem[{Yang et~al.(2019)Yang, Dai, Yang, Carbonell, Salakhutdinov, and
  Le}]{yang2019xlnet}
Yang, Z.; Dai, Z.; Yang, Y.; Carbonell, J.; Salakhutdinov, R.~R.; and Le, Q.~V.
  2019.
\newblock Xlnet: Generalized autoregressive pretraining for language
  understanding.
\newblock In \emph{Advances in neural information processing systems},
  5753--5763.

\bibitem[{Yen et~al.(2017)Yen, Huang, Dai, Ravikumar, Dhillon, and
  Xing}]{yen2017ppdsparse}
Yen, I.~E.; Huang, X.; Dai, W.; Ravikumar, P.; Dhillon, I.; and Xing, E. 2017.
\newblock Ppdsparse: A parallel primal-dual sparse method for extreme
  classification.
\newblock In \emph{Proceedings of the 23rd ACM SIGKDD International Conference
  on Knowledge Discovery and Data Mining}, 545--553.

\bibitem[{Yen et~al.(2016)Yen, Huang, Ravikumar, Zhong, and
  Dhillon}]{yen2016pd}
Yen, I. E.-H.; Huang, X.; Ravikumar, P.; Zhong, K.; and Dhillon, I. 2016.
\newblock Pd-sparse: A primal and dual sparse approach to extreme multiclass
  and multilabel classification.
\newblock In \emph{International Conference on Machine Learning}, 3069--3077.

\bibitem[{You et~al.(2019)You, Zhang, Wang, Dai, Mamitsuka, and
  Zhu}]{you2019attentionxml}
You, R.; Zhang, Z.; Wang, Z.; Dai, S.; Mamitsuka, H.; and Zhu, S. 2019.
\newblock Attentionxml: Label tree-based attention-aware deep model for
  high-performance extreme multi-label text classification.
\newblock In \emph{Advances in Neural Information Processing Systems},
  5820--5830.

\bibitem[{Zubiaga(2012)}]{zubiaga2012enhancing}
Zubiaga, A. 2012.
\newblock Enhancing navigation on wikipedia with social tags.
\newblock \emph{arXiv preprint arXiv:1202.5469} .

\end{thebibliography}

\end{document}